\documentclass{article}

\usepackage{arxiv}

\usepackage[utf8]{inputenc} 
\usepackage[T1]{fontenc}    
\usepackage{hyperref}       
\usepackage{url}            
\usepackage{booktabs}       
\usepackage{amsfonts}       
\usepackage{nicefrac}       
\usepackage{microtype}      
\usepackage{lipsum}         
\usepackage{graphicx}
\usepackage{natbib}
\usepackage{doi}
\usepackage{amsmath}
\usepackage{listings}
\usepackage{xcolor} 
\usepackage{pdflscape}
\usepackage{caption}
\usepackage{capt-of}        
\usepackage{newfloat} 

\definecolor{LightGray}{gray}{0.9}

\lstdefinestyle{jsonstyle}{ backgroundcolor=\color{LightGray}, basicstyle=\ttfamily\footnotesize, breaklines=true, frame=single, numbers=left, numberstyle=\tiny, showstringspaces=false, columns=fullflexible } 
\usepackage{cleveref}       

\usepackage{subcaption}
\usepackage{graphicx}

\newif\ifuniqueAffiliation
\uniqueAffiliationtrue

\ifuniqueAffiliation 
\author{Sergio Ram\'irez-Gallego \\
	Thales Alenia Space Spain\\ 
    Tres Cantos\\
    Spain \\
	\texttt{sergio.ramirez@thalesaleniaspace.com} \\
}
\else
\usepackage{authblk}

\newbox{\orcid}\sbox{\orcid}{\includegraphics[scale=0.06]{orcid.pdf}} 
\author[1]{%
	\href{https://orcid.org/0000-0000-0000-0000}{\usebox{\orcid}\hspace{1mm}David S.~Hippocampus\thanks{\texttt{hippo@cs.cranberry-lemon.edu}}}%
}
\author[1,2]{%
	\href{https://orcid.org/0000-0000-0000-0000}{\usebox{\orcid}\hspace{1mm}Elias D.~Striatum\thanks{\texttt{stariate@ee.mount-sheikh.edu}}}%
}
\affil[1]{Department of Computer Science, Cranberry-Lemon University, Pittsburgh, PA 15213}
\affil[2]{Department of Electrical Engineering, Mount-Sheikh University, Santa Narimana, Levand}
\fi

\renewcommand{\shorttitle}{A Semantic Agent Architecture for Sensor Fusion in Multi-Sensor Space Debris Tracking}

\hypersetup{
pdftitle={A Semantic Agent Architecture for Sensor Fusion in Multi-Sensor Space Debris Tracking},
pdfsubject={q-bio.NC, q-bio.QM},
pdfauthor={Sergio Ram\'irez-Gallego},
pdfkeywords={Agentic AI, Debris Fracking, Sensor Fusion},
}
\title{A Semantic Agent Architecture for Sensor Fusion in Multi-Sensor Space Debris Tracking}

\begin{document}
\maketitle

\begin{abstract} 

Space debris tracking requires robust sensor fusion and data association methods capable of handling sensor degradation, uncertainty, and ambiguous object interactions. This paper presents a semantic multi-agent architecture that combines Large Language Model (LLM) reasoning with classical tracking algorithms for optical-event sensor fusion and adaptive data association. Specialized Optical and Event Agents independently assess observations, estimating reliability, spatial consistency, appearance agreement, and temporal coherence. A Fusion Agent integrates these assessments to generate context-aware observations, association costs, and candidate-selection criteria, while a conventional Hungarian algorithm performs the final deterministic assignment. To evaluate the approach, a text-driven scenario generation methodology is introduced to create challenging multi-sensor tracking situations. Experiments involving optical degradation and ambiguous object crossings show that semantic reasoning reduces association ambiguity, improves cost discrimination, decreases the candidate search space, and preserves object identities when geometric information alone is insufficient. The results demonstrate the feasibility and potential of Agentic AI for explainable and reliability-aware sensor fusion in future autonomous space surveillance systems. 

\end{abstract}


\section{Introduction}
\label{sec:introduction}

Space Situational Awareness (SSA) relies on the accurate detection and tracking of resident space objects to support collision avoidance and ensure the long-term sustainability of space operations~\cite{kessler1978collision,hou2023research}. Modern tracking systems typically combine object detection, sensor fusion, state estimation, and data association to maintain object identities over time~\cite{rakai2022dataassociation}. In this context, optical and event cameras provide complementary information: optical sensors capture appearance and object extent, whereas event cameras offer high temporal resolution and robustness to rapid motion and challenging illumination conditions~\cite{gallego2022event, devulapally2026}. However, fusing observations from both modalities remains challenging~\cite{ralph2022fiesta} under harsh space conditions, such as: sensor degradation, synchronization errors, false detections, sensor disagreement, and ambiguous object interactions.

Classical sensor-fusion and data-association approaches rely primarily on numerical measurements such as positions, confidence scores, and motion estimates~\cite{wei2025fusionreview,rakai2022dataassociation}. While effective and common in well-defined scenarios, they have limited ability to reason about contextual factors explaining why observations disagree or why a particular sensor should be trusted under specific conditions. Successful tracking in complex environments therefore requires not only numerical information but also contextual understanding of sensor conditions, object appearance, motion patterns, and potential failure modes. A further limitation in this topic is the lack of realistic multi-sensor datasets combining optical and event-camera observations representing complex and uncertain scenarios (sensor failures, degradations and multiple occlusions). Consequently, many operational situations of interest for autonomous spacecraft remain difficult to evaluate systematically.

This scarcity problem could have been resolved by modern LLMs, which have demonstrated a remarkable ability to generate detailed and coherent content from natural-language descriptions~\cite{sengar2025generative}, enabling the creation of rich hypothetical scenarios that capture complex interactions, failure modes, and environmental conditions. 
LLMs have also demonstrated their ability to perform contextual reasoning, multi-modal information integration, and decision support in uncertain environments~\cite{hou2025driveagent,rodriguezfernandez2024spacecraft}. Finally, they provide a natural interface between heterogeneous sensors and onboard subsystems by representing observations, quality indicators, and operational constraints using a common language-based abstraction~\cite{karagounis2024autonomous}. Together, these capabilities make LLMs attractive candidates for the generation of representative evaluation scenarios, the support of optical-event sensor fusion and tracking applications, and much more in the space ecosystem.

To exploit these capabilities, this work proposes a semantic multi-agent architecture for optical-event sensor fusion and adaptive data association in space debris tracking. In this architecture, specialized Optical and Event Agents independently interpret modality-specific observations for detection, while a Fusion Agent evaluates sensor agreement and generates adaptive association costs and gating thresholds for tracking. These semantic assessments are then consumed by a conventional Hungarian assignment algorithm~\cite{kuhn1955hungarian}, preserving the robustness and interpretability of classical tracking while introducing a contextual reasoning layer.

The main contributions of this work are summarized as follows:

\begin{itemize} 
    \item A semantic multi-agent architecture for optical-event sensor fusion and track-observation association in space debris tracking, enabling contextual interpretation of heterogeneous observations under sensor degradation, ambiguity, and uncertainty. 
    \item A hybrid semantic-classical framework in which LLM agents generate fusion assessments, candidate sets, and association costs, while a conventional Hungarian algorithm performs the final assignment, preserving the robustness, interpretability, and mathematical guarantees of classical tracking methods. 
    \item A text-driven scenario-generation methodology for creating controlled multi-sensor tracking situations involving degradation, ambiguity, and sensor disagreement, providing a practical mechanism for evaluating complex and infrequent operational conditions in the absence of representative optical-event datasets. 
\end{itemize}

To the best of our knowledge, this is the first work exploring Agentic AI for optical-event sensor fusion and track-observation association in a space-domain context. Therefore, the main focus of this study is to establish feasibility and illustrate the potential of semantic reasoning in sensor fusion and tracking through representative and complex scenarios.

\section{Related Work}
\label{sec:related_work}

Classical multi-object tracking relies on state estimation and data association methods based on distances and confidences~\cite{rakai2022dataassociation}. Methods such as global nearest-neighbor assignment, the Hungarian algorithm, Joint Probabilistic Data Association, and Multiple Hypothesis Tracking provide rigorous solutions, but their effectiveness depends on the quality of the association costs, gates, and sensor models. Consequently, close object crossings, degraded measurements, or unreliable confidence estimates can produce incorrect associations even when the underlying optimization is solved correctly.

Space debris perception research has traditionally focused on optical detection and tracking under challenging backgrounds. Recent learning-based methods improve detection and temporal tracking through feature extraction and synthetic training data~\cite{zhuang2025sdt}. Event cameras have also been investigated for SSA because their high temporal resolution and dynamic range are suitable for faint, fast-moving objects and difficult illumination conditions. Existing approaches demonstrate event-based localization, velocity estimation, and multi-object tracking using asynchronous processing, filtering, learned representations, and multi-stage association~\cite{ralph2022fiesta,zhou2024ebstracker}. However, existing space datasets and methods generally address optical or event sensing separately~\cite{jawaid2026event} (specially for debris tracking), leaving limited support for studying contextual fusion between both modalities under controlled sensor degradation, disagreement, and association ambiguity.

Sensor fusion techniques are commonly categorized according to the stage at which information is combined~\cite{wei2025fusionreview}. Data-level fusion merges raw measurements and retains detailed sensor information, but generally requires accurate calibration and synchronization. Feature-level fusion combines intermediate representations, such as contours, detections, descriptors, or learned embeddings, whereas decision-level fusion integrates independently generated predictions, hypotheses, or confidence assessments. The proposed architecture mainly follows a hybrid decision-level strategy, since the Fusion Agent combines the outputs of specialized sensor agents, while also exploiting feature-level cues such as appearance, motion, and observation geometry.

Finally, LLMs and language-based agents have recently been investigated as contextual reasoning and decision-support components in autonomous systems. \textit{Language Models are Spacecraft Operators} demonstrated that an LLM agent can translate spacecraft-state descriptions into control actions in a simulated environment~\cite{rodriguezfernandez2024spacecraft}, overcoming current state-of-the-art in reinforcement learning. Also, LLM-based autonomous-perception research has recently explored contextual interpretation and heterogeneous sensor integration using text descriptions~\cite{karagounis2024autonomous}. More recent work has considered agentic LLMs for multimodal information fusion and decision support in Space Domain Awareness~\cite{courey2025agentic} at ground level. 

While prior studies have explored LLMs for autonomous operations and decision support, their application to sensor fusion and data association remains largely unexplored. This work investigates how specialized language-based agents can support these tasks while preserving the deterministic behavior of classical tracking algorithms.

\section{Methods}
\label{sec:methods}

Space debris tracking can be formulated as a tracking-by-detection problem in which observations acquired at a given time must be associated with previously estimated object tracks. Let

\begin{equation}
\mathcal{T}=\{T_1,\ldots,T_N\}
\end{equation}

denote the set of predicted tracks and

\begin{equation}
\mathcal{O}=\{O_1,\ldots,O_M\}
\end{equation}

the set of observations provided by one or more sensors. A track represents the estimated state of a physical object, typically including its position, velocity, and additional descriptive information propagated from previous measurements. An observation corresponds to a newly detected object and may contain quantities such as image coordinates, bounding boxes, timestamps, confidence scores, appearance descriptors, or motion signatures.

The objective of data association is to determine the most likely correspondence between tracks and observations. This process is commonly represented by an association cost matrix

\begin{equation}
C =
\begin{bmatrix}
c_{11} & \cdots & c_{1M} \\
\vdots & \ddots & \vdots \\
c_{N1} & \cdots & c_{NM}
\end{bmatrix},
\end{equation}

where each element $c_{ij}$ quantifies the incompatibility between track $T_i$ and observation $O_j$. Lower costs indicate more likely associations. Before assignment, candidate pairs are typically filtered through validation gates that eliminate physically implausible correspondences according to spatial, temporal, or statistical constraints. The remaining track-observation pairs are then evaluated by an assignment algorithm to obtain a one-to-one correspondence between tracks and measurements.


The proposed framework introduces a semantic multi-agent architecture for space debris tracking with sensor fusion. Unlike conventional tracking systems, where all decisions are derived solely from geometric distances and sensor measurements, the proposed approach combines classical estimation algorithms with LLM-based reasoning agents. 


\subsection{Architecture}

\begin{figure}
    \centering
    \includegraphics[width=0.8\linewidth]{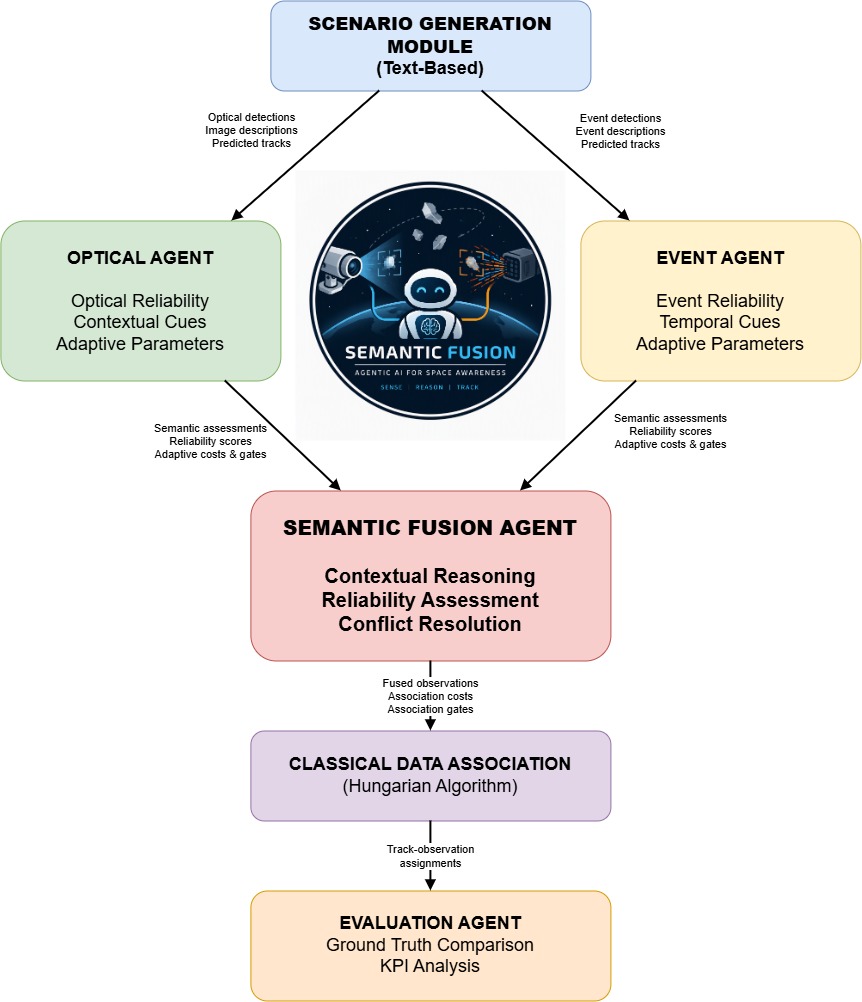}
    \caption{Multi-Agent Architecture for Sensor Fusion and Debris Detection/Tracking. Scenario generation and evaluation agent are sidecar components to be removed in a production environment.}
    \label{fig:architecture}
\end{figure}

The agentic architecture follows a hierarchical decision process in which optical and event-camera observations are first interpreted independently by specialized sensor agents. The simulator provides synthetic debris-tracking scenarios together with predicted object tracks and sensor observations. Sensor-specific agents independently analyze optical and event-camera detections and enrich them with semantic information. The fusion agent reasons over the outputs of both sensors and generates unified observations. Finally, the evaluation module compares the semantically supported association with a classical baseline.
The proposed architecture is composed of five principal modules, illustrated conceptually in Figure~\ref{fig:architecture}.

\subsection{Scenario Generation Module}
\label{sec:scenario_generation}

\begin{figure}[t]
    \centering
    \includegraphics[width=0.99\linewidth]
    {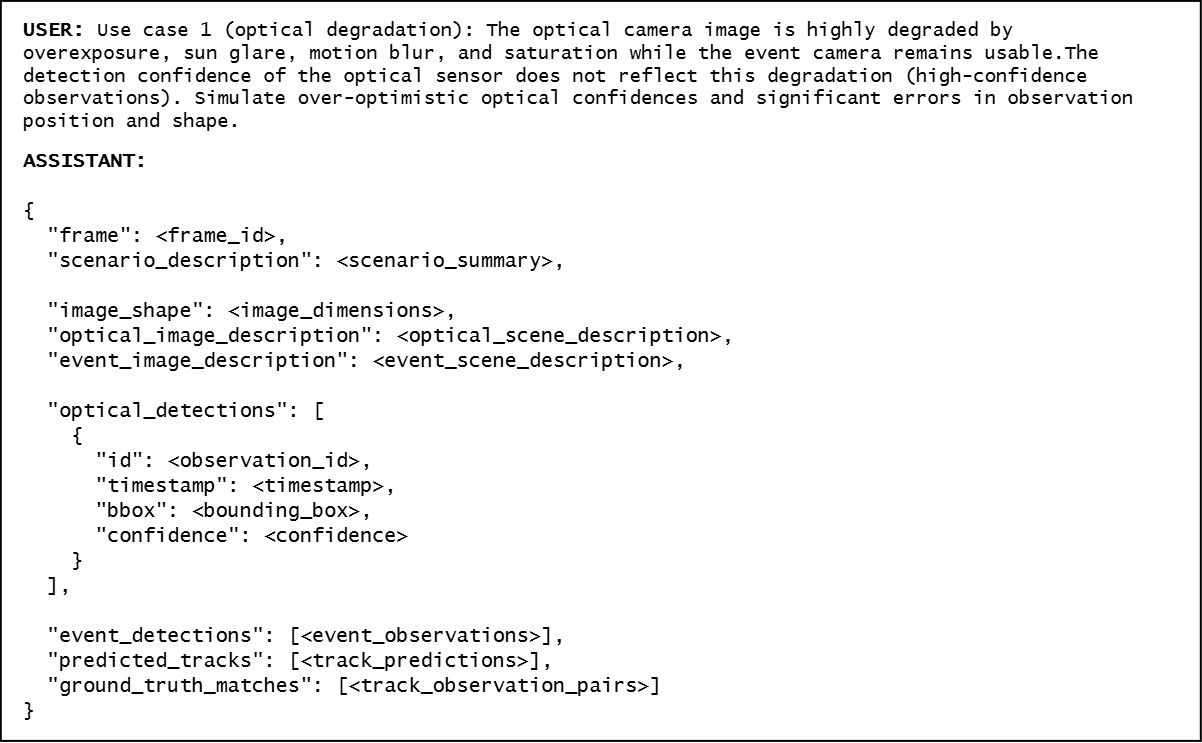}
    \caption{Scenario generation interface. A natural-language use-case description defines the intended sensing conditions, and the generator returns the structured JSON scenario consumed by the multi-agent workflow.}
    \label{fig:scenario-generator}
\end{figure}

The Scenario Generation Module produces the structured input consumed by the Optical, Event, and Fusion Agents. The module receives a natural-language use-case description specifying the intended scene conditions and sources of ambiguity, such as optical degradation, occlusion, asynchronous measurements, sensor disagreement, or closely spaced objects. Based on this description, the generator creates a single-frame synthetic tracking scenario represented as a JSON object. An example of the prompt and output structure is shown in Figure~\ref{fig:scenario-generator}.

Each generated scenario contains four main types of information:

\begin{itemize}
    \item \textbf{Scene metadata:} frame identifier, image dimensions, scenario description, and modality-specific descriptions of the optical and event observations.
    
    \item \textbf{Sensor observations:} optical and event detections containing an observation identifier, timestamp, detector confidence, and bounding box in image coordinates.
    
    \item \textbf{Predicted tracks:} the prior tracking information available before processing the current observations, including a track identifier, timestamp, predicted position, kinematic state, and semantic description of the tracked object.
    
    \item \textbf{Ground-truth associations:} track-observation correspondences used exclusively by the Evaluation Agent to score the association results.
\end{itemize}

The modality-specific descriptions provide scene-level and per-observation context that cannot be represented by bounding boxes and confidence values alone. For the optical channel, this may include visibility, blur, glare, saturation, illumination, occlusion, and boundary quality. For the event channel, the descriptions may represent event density, motion trails, temporal coherence, noise, and dropout. The numerical detections and predicted tracks provide the geometric information required to calculate observation-track distances and construct the association problem. 

In the current prototype, modality-specific textual descriptions replace the semantic information that would normally be extracted from real optical images and event streams (using vision-language models, for instance). In a production environment, real sensor data would instead be processed by conventional or learning-based detectors, incorporated into the Optical and Event Agents as callable tools, to obtain detections, features, confidence scores, and quality indicators. The agents would interpret these outputs and their context, while the Fusion Agent would combine the resulting assessments for data association. Thus, the Scenario Generation Module emulates the upstream sensing and perception pipeline \textit{without} assuming a specific detector implementation.

\subsection{Optical Reasoning Agent}

The Optical Agent receives the optical scene description, detections, image dimensions, and predicted tracks. A geometric distance map is first computed between every detection and track pair. The agent then evaluates optical quality factors such as visibility, illumination, glare, saturation, motion blur, occlusion, trail integrity, and boundary clarity. For each observation-track pair, three semantic consistency metrics are estimated:

\begin{itemize}
    \item \textbf{Spatial consistency}: agreement between the observation and the predicted track position.
    \item \textbf{Appearance consistency}: agreement between the observed visual characteristics and the track description.
    \item \textbf{Timestamp consistency}: temporal coherence between the observation and the predicted track state.
\end{itemize}

These metrics allow the agent to identify situations where geometric and semantic evidence disagree. For example, an observation may be spatially closest to one track while its morphology, appearance, or motion characteristics are more consistent with another. Based on this analysis, the agent estimates observation-level and scene-level reliability, identifies possible failure modes, and generates track-dependent association costs and validation gates. In addition, the agent produces natural-language explanations that justify the generated parameters and preserve potentially ambiguous alternatives for later fusion. A simplified representation of the Optical Agent output is shown at Listing~\ref{lst:optical}.

\begin{lstlisting}[language={}, caption={Optical Agent output format. It includes semantic assessment, parameter estimation and numerical information. Tracks are represented as $T_j$, and observations as $O_i$}, label={lst:optical}]
{
  "distance_map": "{Oi : {Tj : distance}"},
  "observations": [
    {
      "id": "...",
      "semantic_description": "...",
      "semantic": {
        "spatial_consistency": "{Tj : score}",
        "appearance_consistency": "{Tj : score}",
        "timestamp_consistency": "{Tj : score}",
        "reliability": "score"
      },
      "parameters": {
        "association_cost": "{Tj : cost}",
        "association_gate": "{Tj : gate}"
      }
    }
  ],
  "scene_semantic": {
    "reliability": "score",
    "failure_mode": "..."
  }
}
\end{lstlisting}

\subsection{Event-Based Reasoning Agent}

The Event Agent follows the same observation-track evaluation process using event-camera detections and predicted tracks. In addition to spatial, appearance, and timestamp consistency, it evaluates event-specific indicators including event density, event support, noise level, trail integrity, temporal coherence, and sensor dropout. These indicators characterize the quality of the motion evidence and allow the agent to distinguish between reliable object signatures and incomplete or noisy event activity.

The output structure mirrors that of the Optical Agent. As with the optical modality, the Event Agent generates natural-language explanations describing how the observed event patterns support or contradict candidate track associations. These outputs are subsequently consumed by the Fusion Agent, which combines the complementary evidence provided by both sensing modalities. A simplified example of the Event Agent semantic layer output for a single detection is shown at Listing~\ref{lst:event}.

\begin{lstlisting}[language={}, caption={Event Agent semantic layer example. It includes consistency factors, explainable decisions, and other evaluated factors. Tracks are named as $T_j$.}, label={lst:event}]
{
  "semantic": {
    "spatial_consistency": {
      "T1": 0.24,
      "T2": 0.58,
      "T3": 0.42
    },
    "appearance_consistency": {
      "T1": 0.08,
      "T2": 0.91,
      "T3": 0.14
    },
    "timestamp_consistency": {
      "T1": 0.93,
      "T2": 0.93,
      "T3": 0.93
    },
    "trail_integrity": "fragmented",
    "boundary_clarity": "medium",
    "event_density": "low",
    "noise_level": "low",
    "dropout": "partial",
    "event_support": "medium",
    "reliability": 0.68,
    "explainability":
      "The elongated vertical structure and downward-left trail
      agree with T2, while overlap suppresses upper events and
      reduces support. The nearest position is T3, but its broad
      reflective upward-moving foil signature is inconsistent."
  }
}
\end{lstlisting}

\subsection{Semantic Fusion Agent}

The Fusion Agent constitutes the central reasoning component of the architecture. It receives observations and semantic assessments from both sensor agents and determines how information should be combined before association.

Instead of performing purely numerical fusion, the agent evaluates the contextual agreement between sensors and generates a semantically enriched observation representation. The fusion process considers multiple factors, including: sensor reliability scores, spatial consistency, temporal consistency, appearance consistency, sensor agreement, detection confidence, observation freshness, and environmental conditions to assess the reliability of observations and generate adaptive association parameters.

Based on this information, the fusion agent can: merge optical and event observations when both sensors provide mutually consistent evidence, retain only the most reliable sensor when observations are conflicting, reject detections considered unreliable or likely to be false positives. It also generates explanations describing the rationale behind decisions.

Each fused observation contains both numerical and semantic attributes. Numerical attributes include processed object position, bounding box information, confidence values, and track-dependent association parameters. Semantic attributes describe fusion quality, sensor agreement, reliability, and contextual reasoning. Consequently, the fusion stage transforms raw sensor measurements into observations that encode both geometric evidence and contextual understanding. This information is key in complex scenarios where the fused detection will fail deeply, like occlusions or camera degradation. Fusion reasoning for a particular merging candidate (optical observation O2 and event E2) is illustrated at Listing~\ref{lst:fusion}.

\begin{lstlisting}[language={}, caption={Fusion Agent partial output example (most relevant fields for fusion). Tracks are named as $T_j$, Optical observations as $O_i$, and Events as $E_i$}, label={lst:fusion}]
{    
  "sensor_agreement": 0.91,
  "fusion_quality": 0.72,
  "fusion_explanation": "O2 and E2 agree on an elongated vertical strip and downward-left 
  motion. Optical provides the more complete extent but has low boundary clarity from 
  upper-end overlap and haze; event contributes sharper motion direction but has low 
  density, fragmented trail support, and partial overlap-induced dropout. The fused 
  observation is spatially nearer T3, but both sensor semantics consistently favor T2 
  over broad upward-moving reflective foil.",
  "reliability": 0.7
}
\end{lstlisting}

\subsection{Data Association}

Given the set of predicted tracks $\mathcal{T}$ and observations $\mathcal{O}$, the role of the data-association stage is to determine the most likely one-to-one correspondence between them. In conventional multi-object tracking systems, this problem is commonly formulated as a linear assignment problem and solved using the Hungarian algorithm. The proposed architecture preserves this classical formulation. The Optical, Event, and Fusion Agents do not perform the assignment directly; instead, they generate the input information required to construct the candidate set and the corresponding cost matrix.

For each pair, the agents produce an association cost $c_{ij}$ and a validation gate $g_{ij}$, which respectively represent the plausibility of the correspondence and the maximum admissible separation between a track and an observation. Let $d_{ij}$ denote the geometric distance (or agent-derived compatibility score) between track $T_i$ and observation $O_j$. A pair is considered a valid candidate only if

\begin{equation}
d_{ij} \leq g_{ij}.
\end{equation}

The remaining candidate pairs define the association cost matrix

\begin{equation}
C =
\begin{bmatrix}
c_{11} & \cdots & c_{1M} \\
\vdots & \ddots & \vdots \\
c_{N1} & \cdots & c_{NM}
\end{bmatrix},
\end{equation}

which is subsequently processed by the Hungarian algorithm to obtain the globally optimal one-to-one assignment. The optimization problem can be expressed as

\begin{equation}
\min_X \sum_{i=1}^{N}\sum_{j=1}^{M} c_{ij}x_{ij},
\end{equation}

subject to

\begin{equation}
\sum_{j=1}^{M} x_{ij} \leq 1,
\qquad
\sum_{i=1}^{N} x_{ij} \leq 1,
\end{equation}

where $x_{ij}=1$ indicates that observation $O_j$ is assigned to track $T_i$.

The Hungarian algorithm remains responsible for the final deterministic decision, ensuring compatibility with conventional tracking systems and preserving the mathematical guarantees of classical assignment methods.

\subsection{Evaluation Component}

The final stage of the framework evaluates the effect of semantic reasoning on the track-observation association process. Given the ground-truth correspondences generated by the Scenario Generation Module, two association strategies are executed and compared:

\begin{itemize}
    \item \textbf{Baseline association}: candidate pairs are generated using a non-restrictive validation gate and association costs are computed from \textbf{\textit{geometric distance and detector confidence}}.
    
    \item \textbf{Semantic association}: candidate pairs are filtered using the adaptive gates generated by the agents, while association costs are produced by the Fusion agent.
\end{itemize}

For each strategy, the Hungarian algorithm computes the final assignment matrix $X$. The resulting correspondences are then compared against the ground-truth associations to compute performance indicators, such as association accuracy, precision, recall, candidate-space reduction, and cost discrimination. 
These metrics quantify the ability of the semantic layer to reduce ambiguity, improve association selectivity, and generate more informative optimization inputs.

Within the architecture, the evaluation stage can be implemented either as a dedicated Evaluation Agent or as a conventional analysis tool. Unlike the reasoning agents, it does not participate in the association process itself; instead, it provides an objective comparison between classical and semantic-assisted association using the available ground truth. This component is only required for evaluation purposes, and will be removed in production environments. The following section applies this evaluation procedure to two representative debris-tracking scenarios.

\section{Experiments}
\label{sec:experiments}

The objective of the experimental evaluation is to analyze the behavior of the proposed semantic multi-agent architecture during sensor fusion and data association. 
Since the scenarios are synthetically generated and designed to expose specific failure modes, the reported results should be interpreted as a proof-of-concept evaluation of the reasoning framework rather than as a benchmark against state-of-the-art. The objective is twofold: first, to investigate how contextual reasoning influences sensor fusion and data association under ambiguous multi-sensor conditions; and second, to evaluate whether semantic information can generate more selective candidate sets, more discriminative association costs, and more informative decision processes than a purely geometry-based baseline.

\subsection{Experimental Design}

The experimental framework was implemented in Python using LangGraph for multi-agent workflow orchestration and GPT-5.4~\cite{singh2026openaigpt5card} for the semantic reasoning performed by the Optical, Event, Fusion, and Evaluation components. OpenCV was used for geometric calculations, visualization, and generation of the debugging views employed throughout the analysis.

Two representative use cases are considered for study:

\begin{itemize}
    \item The first scenario evaluates severe \textbf{optical degradation}, where the optical sensor suffers from overexposure, glare, motion blur, and saturation while the event sensor remains largely unaffected. 
    \item The second scenario is designed to study \textbf{crossing objects with ambiguous geometry}, where heterogeneous sensor cues provide evidence regarding object identity and localization. 
\end{itemize}

For each scenario, two association modes are compared. The \textbf{\textit{baseline}} uses a conventional Hungarian assignment with large gating thresholds and costs derived from geometric distance and detector confidence. The \textbf{\textit{proposed approach}} uses the same Hungarian optimization algorithm but replaces the fixed association parameters with the adaptive gates and costs generated by the corresponding Agents. Fusion is performed by agents in the same manner for both approaches. No numerical evaluation is performed on fusion performance as reference detections are synthetic. 

\subsubsection{Use Case 1: Optical Degradation}

\begin{figure}[h]
    \centering
    \begin{subfigure}[b]{0.48\textwidth}
        \centering
        \includegraphics[width=\textwidth]{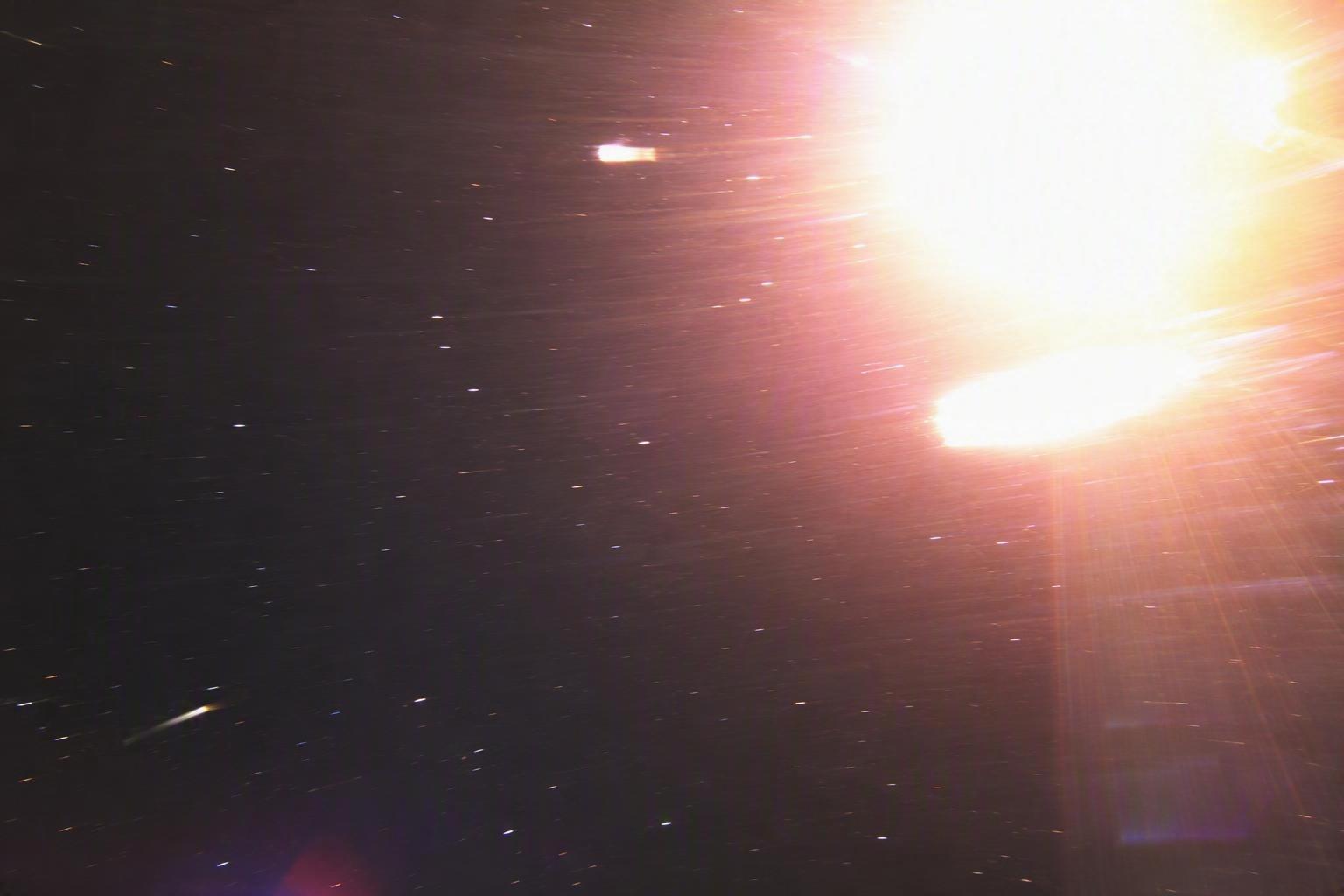}
        \caption{Illustrative optical capture (AI-gen). Sun glare and motion-blurred debris are properly represented. Optical-camera frame depicting four orbital debris fragments (two hidden by light)}
        \label{fig:optical-mockup}
    \end{subfigure}
    \hfill
    \begin{subfigure}[b]{0.48\textwidth}
        \centering
        \includegraphics[width=\textwidth]{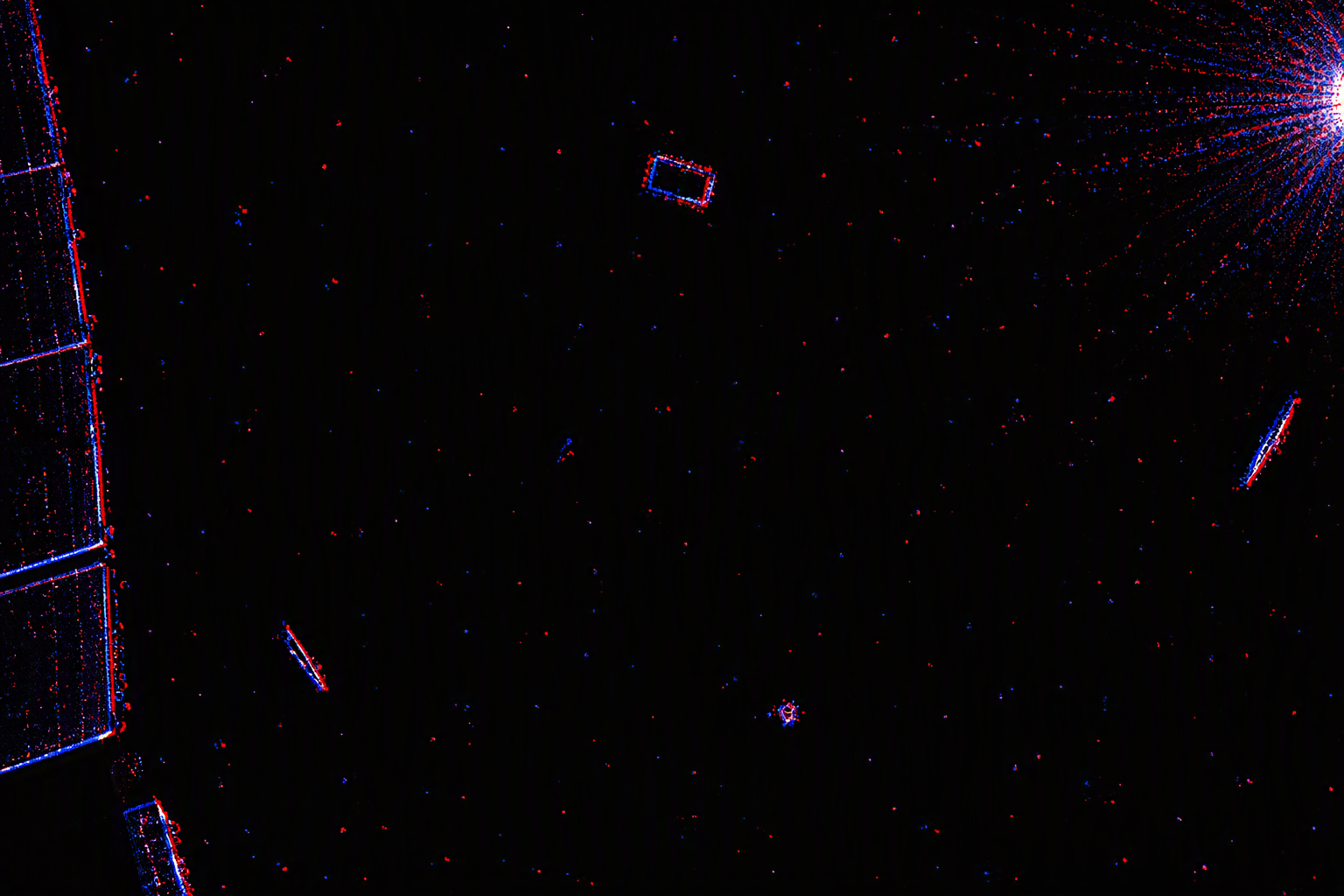}
        \caption{Illustrative event capture (AI-gen). Event-camera frame depicting four orbital debris fragments as sparse red/blue polarity event cluster.}
        \label{fig:event-mockup}
    \end{subfigure}

    \caption{Visual interpretation of Use Case 1 generated from the scenario description. These renderings are provided for qualitative interpretation only and are not used by the association process.}
    \label{fig:case-mockup}
\end{figure}

The first experiment considers a situation in which three known debris objects are observed simultaneously by an optical camera and an event camera. The scene is affected by severe sunward illumination, producing glare, overexposure, bloom, motion blur, and saturation in the optical image. Although the optical detector reports high confidence values (over-optimistic), the resulting detections exhibit significant distortions in object position, and shape. In contrast, the event camera remains largely reliable and provides cleaner localization cues. An AI-generated illustration of this case is depicted at Figure~\ref{fig:case-mockup}. On the other hand, Appendix~\ref{app:usecase1_reasoning} shows a more detailed and accurate (but less realistic) representation of the scene. All figures were generated with OpenCV for qualitative analysis only.

In this scenario, three detections correspond to three existing tracks, while two additional observations represent ambiguous measurements: an optical-only detection generated by glare near the image border and a weak event-only detection corresponding to a potential new object. The scenario is intentionally constructed such that geometric information alone is not sufficient to assess observation quality. In particular, the optical observations associated with tracks T2 and T3 present substantial localization errors despite having confidence values above 0.89.

The Optical Agent correctly identifies the degradation affecting the image and assigns a low scene-level reliability score (0.33). Furthermore, it classifies the frame failure mode as glare and downweights the optical-only artifact despite its high detector confidence. Conversely, the Event Agent reports a scene reliability (0.83) and produces highly confident semantic assessments for the three genuine debris objects. These assessments are subsequently combined by the Fusion Agent, which prioritizes the spatial information provided by the event camera while smartly retaining complementary shape information from the optical sensor. As a result, the fused observations F1-F3 remain consistent with the three predicted tracks, whereas the optical artifact F5 and the weak event cue F4 are assigned low reliability values and treated conservatively.

The main benefit of the proposed approach in this scenario is its ability to account for the different reliability levels of the sensing modalities. The Optical Agent identifies that the upper debris object is negatively affected, while the Event Agent provides cleaner motion and localization evidence. Rather than combining both measurements uniformly (standard fusion), the Fusion Agent gives greater importance to the event-based observation for localization while retaining the useful appearance information available from the optical sensor. As a result, the fused detections exhibit more accurate object positions, dimensions, and shape descriptions than those obtained from the sensors alone or from a simple fusion strategy. At the association level, the semantic agents generate more selective candidates and representative costs. 

\subsubsection{Use Case 2: Crossing Objects with Ambiguous Geometry}

\begin{figure}[h]
    \centering
    \begin{subfigure}[b]{0.48\textwidth}
        \centering
        \includegraphics[width=\textwidth]{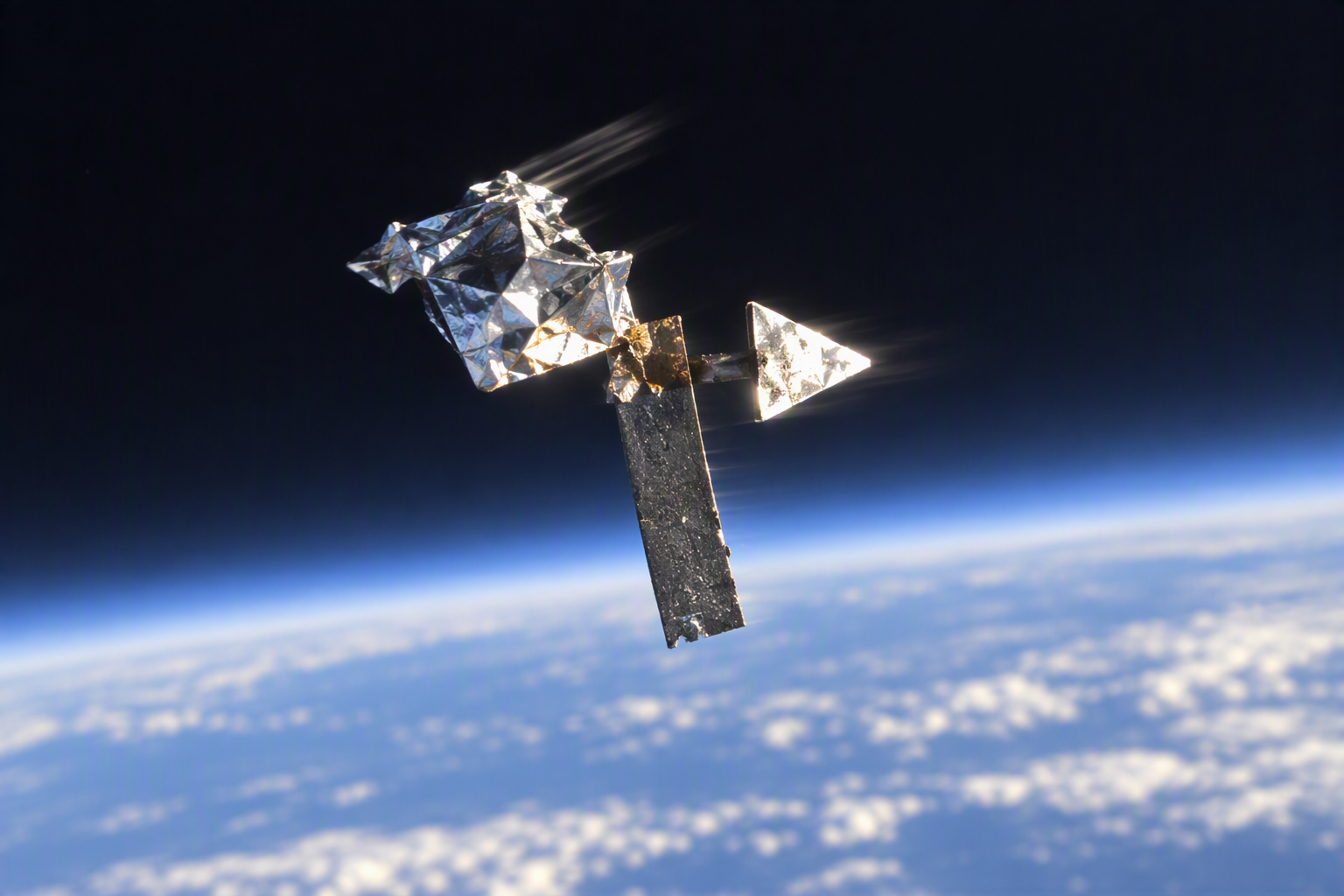}
        \caption{Illustrative optical capture (AI-gen). Zoomed optical image of a debris-crossing scenario observed by a satellite camera. Three closely spaced debris objects exhibit partial occlusion.}
        \label{fig:optical-crossing}
    \end{subfigure}
    \hfill
    \begin{subfigure}[b]{0.48\textwidth}
        \centering
        \includegraphics[width=\textwidth]{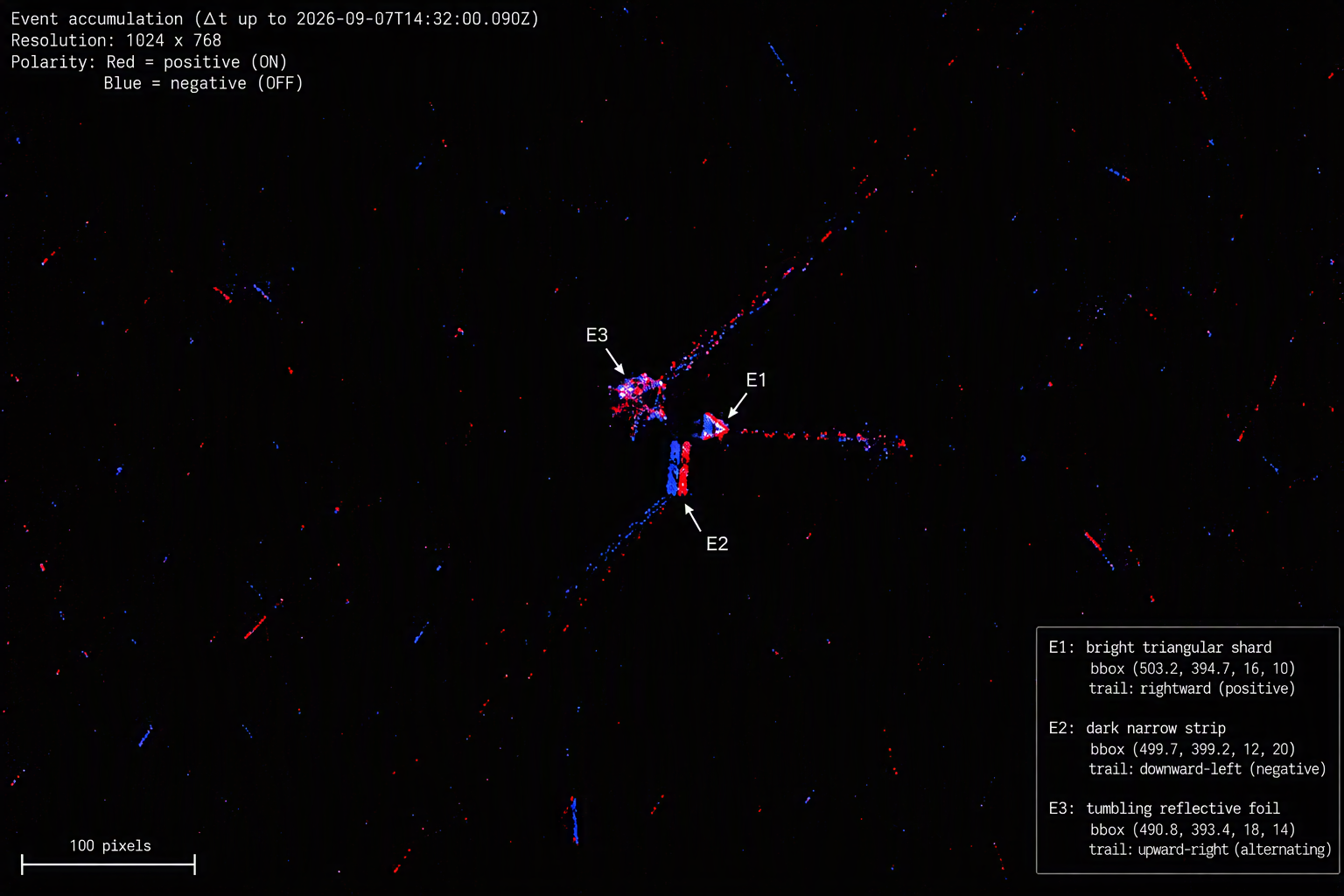}
        \caption{Illustrative event capture (AI-gen). Event accumulation image from a event camera showing three debris objects crossing. Motion-induced positive and negative polarity is depicted.}
        \label{fig:event-crossing}
    \end{subfigure}

    \caption{Visual interpretation of Use Case 2 generated from the scenario description. This scene represents a challenging detection and tracking case. Both subfigures represents the same scene with different zoom levels. These renderings are provided for qualitative interpretation only and are not used by the association process.}
    \label{fig:crossing-mockup}
\end{figure}

The second experiment evaluates a challenging crossing scenario in which three debris objects occupy nearly identical positions while maintaining distinct appearance and motion characteristics. The scenario is deliberately constructed such that each observation is geometrically closer to an incorrect track than to its true correspondence, creating a cyclic ambiguity that cannot be resolved reliably using spatial information alone. In this case, successful association requires exploiting semantic cues such as shape, brightness, contour, and motion patterns. An AI-generated illustration of this case is depicted at Figure~\ref{fig:crossing-mockup}.

The Optical and Event Agents independently identify the characteristic signatures of the three objects, namely a bright triangular shard (T1), a dark elongated strip (T2), and an irregular reflective foil fragment (T3). During fusion, both modalities agree on these semantic identities despite the misleading geometric proximity. Consequently, the Fusion Agent generates association costs that strongly favor appearance-consistent matches while maintaining crossing-tolerant gating.



The semantic agents exploit complementary appearance and motion information to distinguish between the objects despite their overlapping trajectories. In particular, the Optical Agent identifies shape- and morphology-based cues, while the Event Agent provides \textbf{\textit{distinctive motion}} signatures that remain consistent throughout the crossing. The Fusion Agent combines both sources of evidence to preserve the identity of each object and generate fused detections with more coherent descriptions than those produced by geometric or simpler fusion approaches. 

From the association perspective, the system operates on a representation in which distinct objects remain distinguishable despite their proximity. The baseline assignment, which relies primarily on geometric proximity and detector confidence, produces incorrect identity swaps. In contrast, the semantic fusion approach preserves object identities and increases association accuracy, and cost/gates optimization. 

\subsubsection{Multi-case association evaluation}

Focusing on the numerical evaluation of data association, Table~\ref{tab:association_results} compares both approaches using four complementary metrics. The number of valid candidate pairs and the candidate reduction quantify the ability of the semantic layer to prune unlikely associations before optimization. Association accuracy and precision evaluate correspondence correctness with respect to the ground truth, while the mean correct match cost measures how strongly the generated cost matrix favors correct assignments. Lower costs indicate more discriminative association hypotheses. The cost reduction metric compares the average cost assigned to ground-truth associations by the semantic and baseline approaches.

The results show that the proposed semantic layer produces more informative and discriminative costs than the baseline approach. In the optical degradation scenario, both methods achieve perfect association accuracy, but semantic reasoning reduces the number of valid candidates by 66.7\% and lowers the mean cost of correct matches by 75.5\%. This indicates that the semantic agents successfully remove ambiguous alternatives while preserving all correct correspondences. In the crossing-objects scenario, where geometric proximity is intentionally misleading, the semantic agents exploit appearance and motion cues, increasing association accuracy from 33.3\% to 66.7\% while achieving perfect precision. At the same time, the semantic layer reduces the candidate search space by 44.4\% and decreases the mean cost of correct associations by more than 97\%. Notably, the semantic approach avoids the identity swaps produced by the baseline and generates a significantly cleaner optimization problem for the Hungarian assignment algorithm. Gates are set intentionally high for the baseline, and no gate comparison is performed as each method operates in different units.

\begin{table}[t]
\centering
\caption{Comparison of the baseline association and the proposed semantic fusion approach.}
\label{tab:association_results}
\begin{tabular}{lcccc}
\toprule
\textbf{Metric} &
\multicolumn{2}{c}{\textbf{Optical Degradation}} &
\multicolumn{2}{c}{\textbf{Crossing Objects}} \\
\cmidrule(lr){2-3}
\cmidrule(lr){4-5}
&
\textbf{Baseline} &
\textbf{Semantic} &
\textbf{Baseline} &
\textbf{Semantic} \\
\midrule

Valid Candidate Pairs & 9 & 3 & 9 & 5 \\

Candidate Reduction (\%) & -- & 66.7 & -- & 44.4 \\


Association Accuracy & 1.000 & 1.000 & 0.333 & 0.667 \\

Association Precision & 1.000 & 1.000 & 0.333 & 1.000 \\

Mean Correct Match Cost & 0.449 & 0.110 & 1.000 & 0.023 \\

Cost Reduction (\%) & -- & 75.5 & -- & 97.7 \\

\bottomrule
\end{tabular}
\end{table}


\section{Discussion}
\label{sec:discussion}

The experiments demonstrate that semantic agents can generate adaptive association costs and gates that incorporate valuable information about sensor reliability, observation quality, appearance, and motion consistency. 
Empirical outcomes also show that the proposed architecture contributes not only to data association but also to the sensor-fusion process itself. 
Furthermore, the text-driven scenario generation module also proved valuable for constructing controlled multi-sensor tracking situations with specific sources of uncertainty, enabling systematic analysis of sensor degradation, ambiguity, and cross-sensor disagreement without requiring large-scale annotated datasets (now inexistent and hard to generate). Beyond the generated use cases, \textbf{potential applications/scenarios} of the proposed architecture include: sensor complementarity, sensor failure backup, asynchronous sensor fusion, conflicting sensor measurements, and ambiguous object interactions, among others.


Concerning LLM experiments, the rapid evolution of foundation models implies that metrics such as inference speed or reasoning performance may quickly become outdated~\cite{sengar2025generative}. Consequently, the primary contribution of this work is the architectural concept rather than a specific onboard implementation. Similarly, although response time remains an important consideration for spacecraft applications, semantic reasoning does not necessarily need to operate continuously. Agent execution may be triggered only when specific conditions are detected, or integrated within hybrid online-offline frameworks as in GUIDE~\cite{carrasco2026guide}.

Compared with rule-based approaches and task-specific machine learning models, Agentic AI offers greater flexibility, explainability, and adaptability to new situations~\cite{rodriguezfernandez2024spacecraft}. More importantly, the same AI motor could potentially address multiple onboard functions, including sensor fusion, resource management, anomaly interpretation, and mission planning simultaneously. Although this vision remains a long-term objective, it highlights the potential of language-based systems as a common reasoning layer for future autonomous spacecraft, thus reducing subsystem development complexity and effort.

In fact, a primary advantage of LLM-based systems is their use of natural language as a common representation layer, enabling heterogeneous sensors, algorithms, and spacecraft subsystems to exchange contextual information through an unified interface~\cite{karagounis2024autonomous,courey2025agentic}. Such an approach could ultimately reduce the need for developing and maintaining independent decision logic for each subsystem, enabling a more unified and adaptive onboard software architecture.




\section{Conclusions and Future Work}
\label{sec:conclusions}

This paper presented a semantic multi-agent architecture for optical-event sensor fusion and adaptive data association in space debris tracking. The proposed approach combines the contextual reasoning capabilities of LLM agents with the robustness of classical tracking algorithms, allowing sensor reliability, observation quality, and scene context to influence both the fusion process and data association while preserving a deterministic assignment process. In addition, a text-driven scenario-generation methodology was introduced to create controlled multi-sensor tracking situations involving uncertainty, sensor degradation, and association ambiguity. The experimental results demonstrate the feasibility of using Agentic AI to support multi-sensor perception and tracking under uncertainty.

The main contribution of the proposed architecture is the introduction of capabilities that are difficult to achieve using numerical representations alone, including explainable decision-making, adaptive association parameters, reliability-aware sensor fusion, and the integration of heterogeneous information such as geometry, appearance, motion, temporal consistency, and environmental context. In the scenarios evaluated, these capabilities generated more informative and discriminative outputs while preserving the mathematical guaranties of conventional optimization methods.

Although this work should be regarded as an initial proof of concept, the results suggest that Agentic AI represents a promising direction for future space perception systems. Future work will focus on deeper validation: new uncertain/ambiguous image-based scenarios, multi-frame reasoning, asynchronous sensor fusion, and lightweight onboard deployment strategies. 

\bibliographystyle{unsrtnat}
\bibliography{library}

\appendix

\section{Multi-Agent Reasoning Outputs for Use Case 1}
\label{app:usecase1_reasoning}

The following figures provide a detailed view of the information exchanged between the Optical, Event, and Fusion Agents during the processing of the optical-degradation scenario presented in Section~\ref{sec:experiments}. In addition to detections, track predictions, and confidence values, the visualizations expose the intermediate semantic representations generated by the architecture. The upper information panels summarize scene-level reasoning, while the lower panels provide observation-level assessments and generated association parameters. Some labels are too long to be displayed entirely. These views illustrate how semantic information propagates through the system before construction of the association cost matrix and execution of the Hungarian assignment algorithm. 

\begin{landscape}
\begin{figure}[p]
    \centering
    \includegraphics[
        width=0.95\linewidth,
        keepaspectratio
    ]{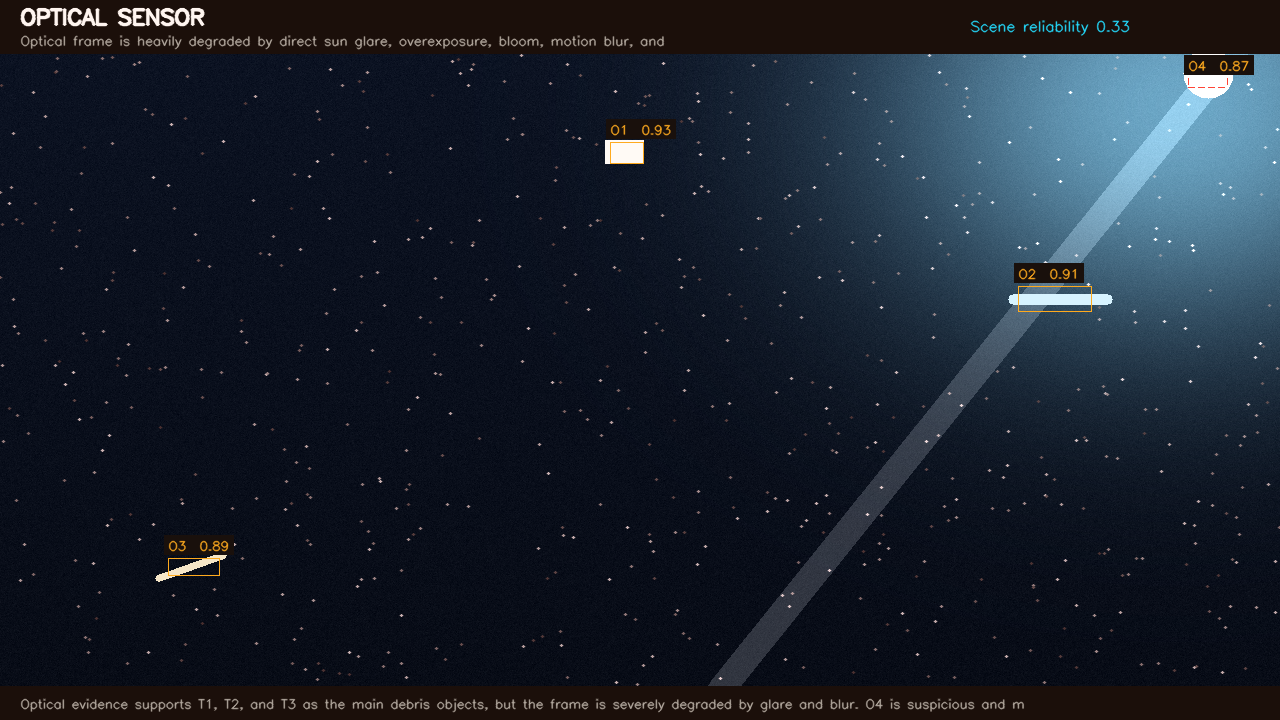}
    \caption{Optical Agent input/output for Use Case~1 (optical degradation). Same scene as Figure~\ref{fig:optical-mockup}. Optical detections with confidences, and semantic assessments are shown.}
    \label{fig:optical-io}
\end{figure}
\end{landscape}

\clearpage

\begin{landscape}
\begin{figure}[p]
    \centering
    \includegraphics[
        width=0.95\linewidth,
        keepaspectratio
    ]{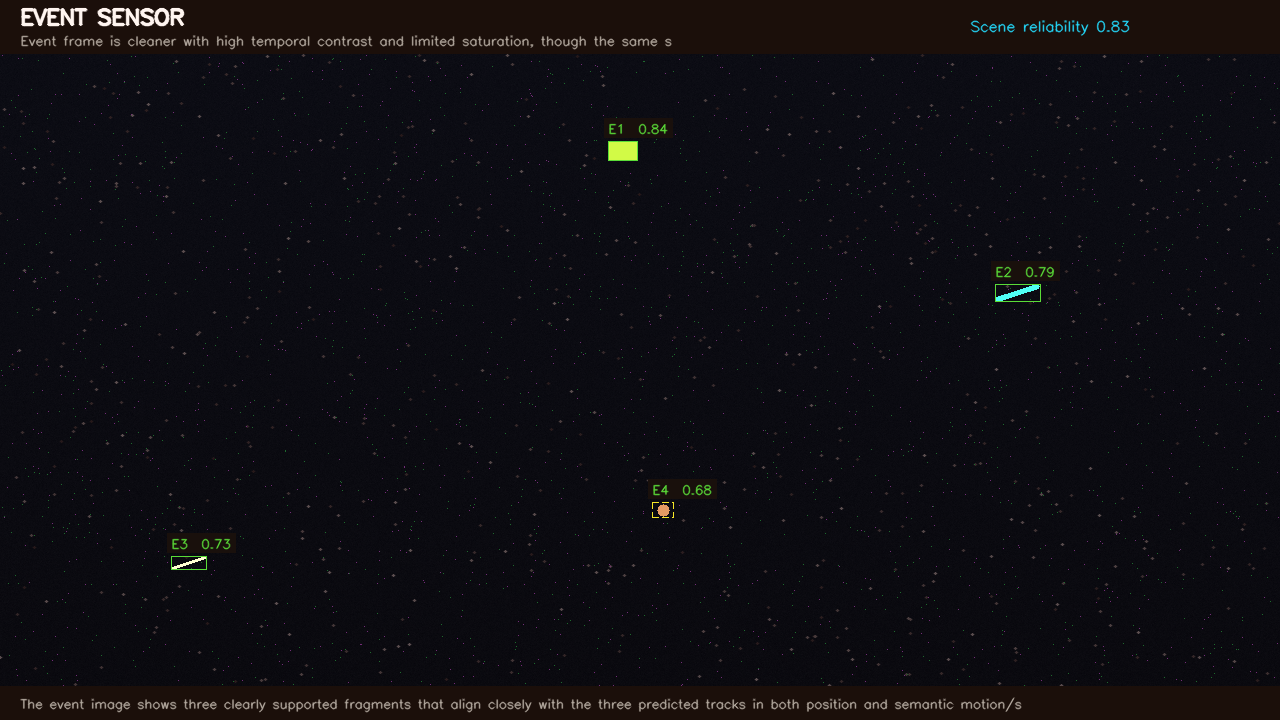}
    \caption{Event Agent input/output for Use Case~1. Same scene as Figure~\ref{fig:event-mockup}. Event-camera detections with confidences, and semantic evaluations are displayed.}
    \label{fig:event-io}
\end{figure}
\end{landscape}

\clearpage

\begin{landscape}
\begin{figure}[p]
    \centering
    \includegraphics[
        width=0.95\linewidth,
        keepaspectratio
    ]{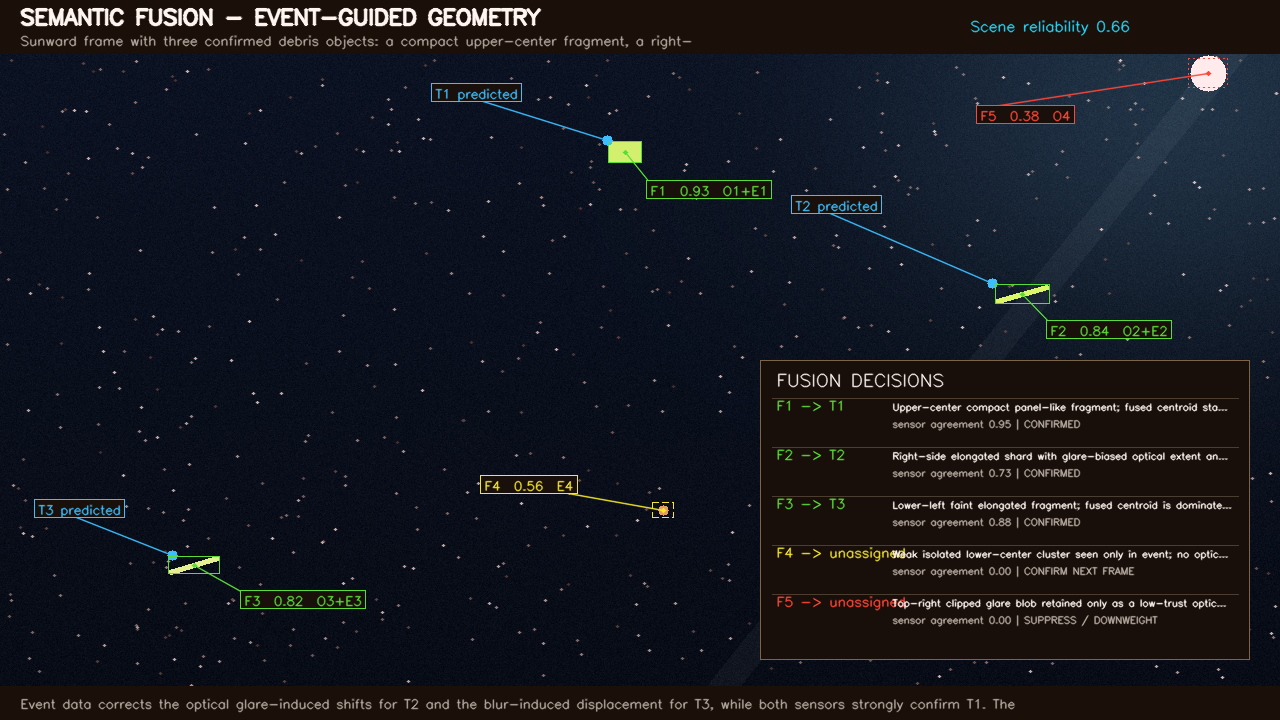}
    \caption{Fusion Agent input/output for Use Case~1. Fusion decisions, fused observations, track positions, confidence values, and semantic explanations are shown.}
    \label{fig:fusion-io}
\end{figure}
\end{landscape}

\end{document}